\documentclass[letterpaper]{article} % DO NOT CHANGE THIS
\usepackage{aaai2026}  % DO NOT CHANGE THIS
\usepackage{times}  % DO NOT CHANGE THIS
\usepackage{helvet}  % DO NOT CHANGE THIS
\usepackage{courier}  % DO NOT CHANGE THIS
\usepackage[hyphens]{url}  % DO NOT CHANGE THIS
\usepackage{graphicx} % DO NOT CHANGE THIS
\usepackage{natbib}  % DO NOT CHANGE THIS AND DO NOT ADD ANY OPTIONS TO IT
\usepackage{caption} % DO NOT CHANGE THIS AND DO NOT ADD ANY OPTIONS TO IT
\usepackage{algorithm}
\usepackage{algorithmic}
\usepackage{amsmath,amssymb,amsfonts}
\usepackage{multirow}
\usepackage{diagbox}
\usepackage{booktabs}
\usepackage{xcolor}

\usepackage{newfloat}
\usepackage{listings}
\DeclareCaptionStyle{ruled}{labelfont=normalfont,labelsep=colon,strut=off} % DO NOT CHANGE THIS
\floatstyle{ruled}
\newfloat{listing}{tb}{lst}{}
\floatname{listing}{Listing}
\title{SynMatch: Rethinking Consistency in Medical Image Segmentation with Sparse Annotations}
\author {
    Zhiqiang Shen\textsuperscript{\rm 1,\rm 2},
    Peng Cao\textsuperscript{\rm 1,\rm 2}
    Xiaoli Liu\textsuperscript{\rm 3}
    Jinzhu Yang\textsuperscript{\rm 1,\rm 2}
    Osmar R. Zaiane\textsuperscript{\rm 4}
}
\affiliations {
    \textsuperscript{\rm 1}School of Computer Science and Engineering, Northeastern University, Shenyang 110819, China\\
    \textsuperscript{\rm 2}Key Laboratory of Intelligent Computing in Medical Image, Ministry of Education, Northeastern University, Shenyang 110819, China\\
    \textsuperscript{\rm 3}AiShiWeiLai AI Research, Beijing, China\\
    \textsuperscript{\rm 4}Alberta Machine Intelligence Institute, University of Alberta, Edmonton, Alberta, Canada\\
}

\begin{document}

\maketitle

\begin{abstract}
Label scarcity remains a major challenge in deep learning-based medical image segmentation.
Recent studies use strong-weak pseudo supervision to leverage unlabeled data.
However, performance is often hindered by inconsistencies between pseudo labels and their corresponding unlabeled images.
In this work, we propose \textbf{SynMatch}, a novel framework that sidesteps the need for improving pseudo labels by synthesizing images to match them instead.
Specifically, SynMatch synthesizes images using texture and shape features extracted from the same segmentation model that generates the corresponding pseudo labels for unlabeled images.  
This design enables the generation of highly consistent synthesized image–pseudo-label pairs without requiring any training parameters for image synthesis.
We extensively evaluate SynMatch across diverse medical image segmentation tasks under semi-supervised learning (SSL), weakly-supervised learning (WSL), and barely-supervised learning (BSL) settings with increasingly limited annotations.
The results demonstrate that SynMatch achieves superior performance, especially in the most challenging BSL setting. 
For example, it outperforms the recent strong-weak pseudo supervision-based method by 29.71\% and 10.05\% on the polyp segmentation task with 5\% and 10\% scribble annotations, respectively.
The code will be released at https://github.com/Senyh/SynMatch.
\end{abstract}

\section{Introduction}
Medical image segmentation for delineating organs, tissues, and lesions is a critical task in computer-aided diagnosis. Deep learning techniques have significantly advanced the progress in this field, relying on large-scale labeled data to train robust segmentation models~\cite{ronneberger2015u,milletari2016v,cao2022swin,wang2022uctransnet,wu2024medsegdiff,wu2024medsegdiffv2}. 
However, dense annotations for medical images are time-consuming and expertise-intensive, requiring radiologists to label target structures at the pixel level~\cite{yu2019uamt,luo2021semi,shen2023co}. 
Learning with limited annotations is crucial for reducing the annotation burden.

%--------------------------------------------------------------------------------------------------------------
\begin{figure}[t]
\centering
\includegraphics[width=\linewidth]{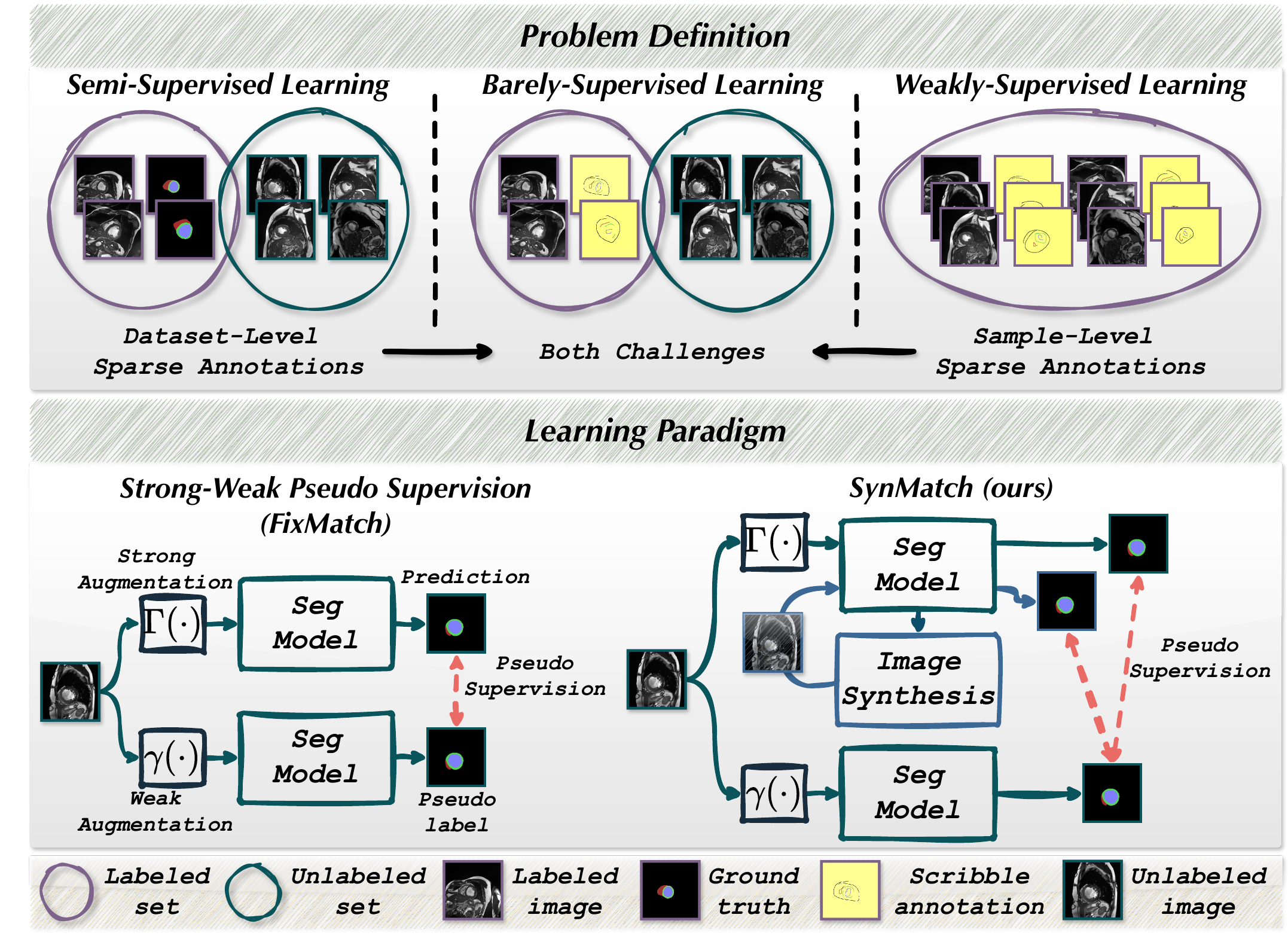} 
\caption{Illustration of (a) the problem definitions for semi-supervised learning, weakly-supervised learning, and barely-supervised learning in medical image segmentation, and (b) the learning paradigms of strong-weak pseudo supervision (FixMatch) and our SynMatch.}
\label{fig:synmatch_intro}
\end{figure}
%--------------------------------------------------------------------------------------------------------------

%--------------------------------------------------------------------------------------------------------------
\begin{figure}[t]
\centering
\includegraphics[width=\linewidth]{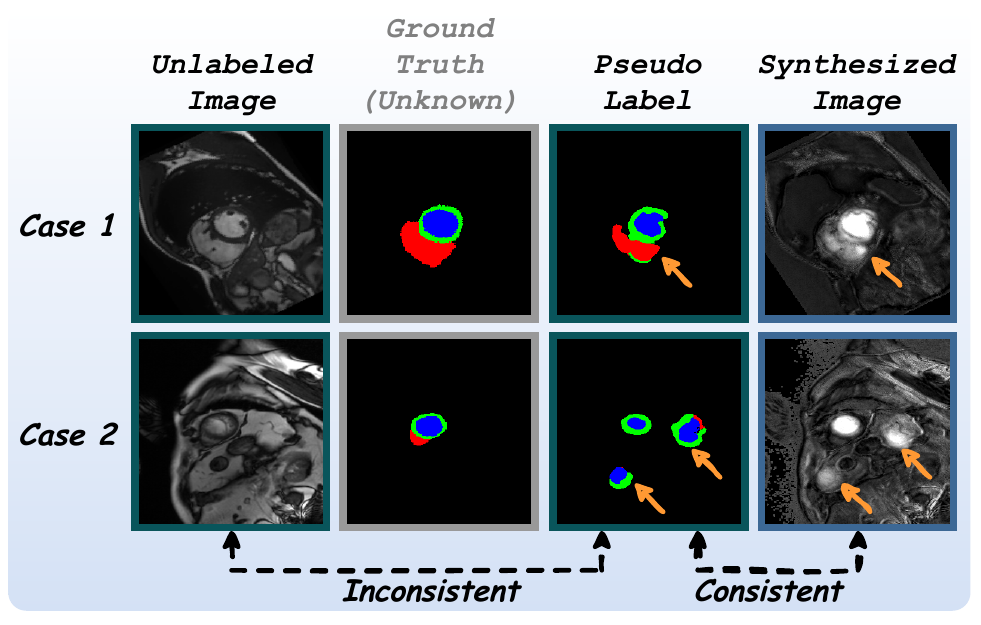} 
\caption{Illustration of inconsistencies between unlabeled images and their corresponding pseudo labels. SynMatch synthesizes new images that align more closely with the pseudo labels. Orange arrows highlight regions where the pseudo labels deviate from the ground truth of the unlabeled images; SynMatch synthesized images effectively correct these inconsistencies, resulting in highly consistent synthesized-image and pseudo-label pairs.}
\label{fig:synmatch_intro_inconsistent_example}
\end{figure}
%--------------------------------------------------------------------------------------------------------------

Semi-supervised learning (SSL)~\cite{wu2021semi,wu2022exploring,wang2023mcf,chi2024adaptive,yang2025semi,li2025boost} and weakly supervised learning (WSL)~\cite{lee2020scribble2label,valvano2021learning,luo2022scribble,zhang2022cyclemix,wei2023weakpolyp,han2024dmsps} have been proposed to enable model training with sparse annotations. As illustrated in Figure~\ref{fig:synmatch_intro}, SSL assumes that the training set consists of a limited number of fully annotated samples and numerous unlabeled images, which can be referred to as dataset-level sparse annotations. 
In contrast, WSL trains a segmentation model using a sample-level, sparse-annotated dataset, \emph{e.g.}, points~\cite{qu2019weakly,qu2020weakly}, bounding boxes~\cite{wei2023weakpolyp}, and scribbles~\cite{valvano2021learning,zhang2022shapepu}, where scribbles provide better delineation for capturing complex anatomical structures than other sparse labeling forms~\cite{luo2022scribble}. 
Going a step further, a more challenging sparse annotation setting includes both a sample-level sparsely labeled set and an unlabeled set [Figure~\ref{fig:synmatch_intro}]. According to definitions in existing literature~\cite{bitarafan20203d,li2022pln,lucas2022barely,cai2023orthogonal,wu2023compete,su2024self}, we define this as a barely-supervised learning (BSL) problem.

Strong-weak pseudo supervision has emerged as a dominant paradigm for addressing medical image segmentation with sparse annotations~\cite{sohn2020fixmatch}.
It underpins many state-of-the-art approaches~\cite{chen2021semi,zhang2022cyclemix,yang2023revisiting,su2024self,yang2025unimatch}.
Specifically, built upon the idea that combines consistency regularization~\cite{sajjadi2016regularization,tarvainen2017mean} and pseudo-labeling~\cite{lee2013pseudo}, strong-weak pseudo supervision enforces consistency between the model's predictions for strongly perturbed unlabeled images and the pseudo labels for weakly perturbed unlabeled images, with a confidence thresholding to improve the quality of pseudo labels [Figure~\ref{fig:synmatch_intro}].
We investigate and rethink the strong-weak pseudo supervision in the more challenging context of medical image segmentation and observe a critical issue: inconsistencies arise between pseudo labels and unlabeled images, regardless of efforts to improve pseudo label quality, as shown in Figure~\ref{fig:synmatch_intro_inconsistent_example}. 
These inconsistencies fundamentally degrade model performance during training.

To address this issue, rather than deliberately pursuing high-quality pseudo labels, we propose a novel perspective: \textit{synthesizing images to match pseudo labels}, enabling the model to be trained on highly consistent synthesized image-pseudo label pairs [Figure~\ref{fig:synmatch_intro_inconsistent_example}]. 
Therefore, we pose a key question: \textit{How can image synthesis be performed to ensure consistency with pseudo-labels?}
In recent years, numerous semantic-to-image methods have been developed to synthesize images conditioned on segmentation maps~\cite{huo2018synseg,park2019semantic,zhu2020sean,lyu2022pseudo,zhu2022label,billot2023synthseg,mayr2024narrowing}. 
However, these methods rely either on large-scale labeled datasets~\cite{zhu2022label,billot2023synthseg} or on task-specific region-of-interest (ROI) masks (\emph{e.g.}, lung lobes)~\cite{lyu2022pseudo} to train generative models with numerous parameters for producing fine-grained and high-quality images. These limitations restrict their applicability to medical image segmentation with sparse annotations, where labeled data are scarce and obtaining ROI masks is labor-intensive and even unavailable. 
To this end, we propose to synthesize images using texture and shape features extracted from the same segmentation model that generates the corresponding pseudo labels for unlabeled images. 
This idea offers several advantages: \textbf{1) Parameter-free}. It requires no additional training parameters for image synthesis, avoiding dependence on large-scale training datasets.
\textbf{2) Semantic Consistency}. Since both the synthesized image and its corresponding pseudo label originate from the same segmentation model (for an unlabeled image), they inherently share unified semantic information. As a result, the pseudo labels can be regarded as the ground truth for the synthesized images [Figure~\ref{fig:synmatch_intro_inconsistent_example}].
\textbf{3) Image Completeness}. The synthesized images preserve essential textures and anatomical structures, enabling them to act as realistic images.
Building upon this idea, we design a novel framework (\textbf{SynMatch}) for medical image segmentation with sparse annotations [Figure~\ref{fig:synmatch_intro}]. 

Our contributions are summarized as follows:
\begin{itemize}
    \item We identify inconsistencies between unlabeled images and pseudo labels in the fundamental strong-weak pseudo supervision paradigm and validate the positive correlation between semantic consistency and segmentation performance.
    \item We propose SynMatch, a generic framework for medical image segmentation with sparse annotations. At the core of our approach is generating highly consistent synthesized-image and pseudo-label pairs, which are leveraged to conduct a powerful complement to strong-weak pseudo supervision. The key idea of synthesizing images to match pseudo labels is simple yet effective, potential to act as a flexible baseline that can be seamlessly integrated with advanced pseudo supervision techniques.
    \item 
    We establish a new benchmark using multiple public datasets with varying levels of label scarcity, including SSL, WSL, and BSL settings, to facilitate comprehensive and robust evaluation of label-efficient medical image segmentation algorithms.
    Experimental results demonstrate the versatility of SynMatch in these settings and its superiority over the state-of-the-art. The results also shed light on the importance of consistency between unlabeled images and pseudo labels for the underlying tasks.
\end{itemize}

%--------------------------------------------------------------------------------------------------------------
\begin{figure*}[t]
\centering
\includegraphics[width=\linewidth]{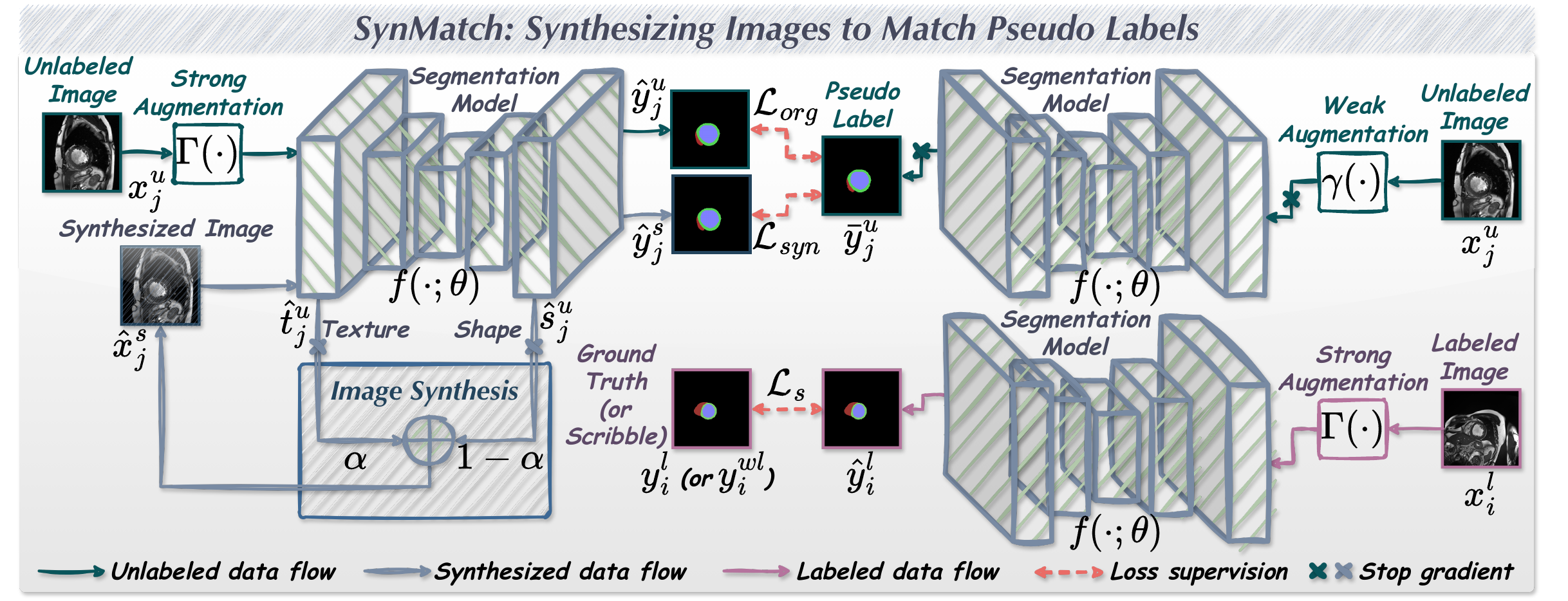} 
\caption{Illustration of the SynMatch framework for medical image segmentation with sparse annotations. SynMatch synthesizes an image $\hat{x}^s_j$ to match the pseudo label $\bar{y}^u_j$ using the texture feature $\hat{t}^u_j$ and shape feature $\hat{s}^u_j$ extracted from the segmentation process of an unlabeled image $x^u_j$. The overall loss supervision includes the supervised learning loss on labeled data and the unsupervised learning losses on both unlabeled images and synthesized images.}
\label{fig:synmatch_framework}
\end{figure*}
%--------------------------------------------------------------------------------------------------------------

\section{Related Work}
Label scarcity poses a critical challenge in medical image segmentation, highlighting the importance of learning from sparse annotations to alleviate the annotation burden. Research in this field can generally be categorized into three distinct settings based on the form of sparse annotation used: 
1) semi-supervised learning~\cite{luo2021semi,wu2021semi,wu2022mutual,shen2023co,wang2023mcf,bai2023bidirectional,chi2024adaptive,kumari2025leveraging,li2025boost,pan2025dusss,zeng2025pick,yang2025semi,li2025dynamic,huang2025gapmatch} under dataset-level sparse annotations, 
2) weakly supervised learning~\cite{qu2019weakly,qu2020weakly,lee2020scribble2label,valvano2021learning,luo2022scribble,zhang2022cyclemix,wei2023weakpolyp,li2023scribblevc,han2024dmsps} under sample-level sparse annotations, 
and 3) barely supervised learning~\cite{bitarafan20203d,li2022pln,lucas2022barely,cai2023orthogonal,wu2023compete,su2024self}: To the best of our knowledge, there is currently no consensus or formal definition for the BSL setting in the existing literature. In this work, we define BSL as a setting combining both SSL and WSL, where models are trained using both dataset-level and sample-level sparse annotations.

The mainstream methodology for these problems follows the strong-weak pseudo supervision paradigm~\cite{sohn2020fixmatch}, which enforces consistency between the model’s predictions for strongly and weakly perturbed unlabeled images. Specifically, recent studies build upon this paradigm to leverage unlabeled data, incorporating elaborately designed components to further enhance learning efficacy, \emph{e.g.}, effective consistency constraints~\cite{luo2021semi,pan2025dusss,yang2025semi,li2025dynamic,huang2025gapmatch,han2024dmsps}, various strong augmentation~\cite{zhang2022cyclemix,shen2023co,chi2024adaptive}, and pseudo label generation and selection strategies~\cite{li2022pln,su2024self,zeng2025pick}. 
However, there is no unified method that performs well across all three settings. To bridge this gap, we propose a generic framework, SynMatch, which synthesizes images to match pseudo labels and achieves superior performance in all three settings.

\section{Methodology}
We first define the SSL, WSL, and BSL settings in medical image segmentation with sparse annotations as follows.
\begin{itemize}
    \item SSL: The training set $\mathcal{D} = \{\mathcal{D}^L, \mathcal{D}^U\}$ consists of a labeled set $\mathcal{D}^L = \{(x^l_i, y^l_i)_{i=1}^{N^L}\}$ and an unlabeled set $\mathcal{D}^U = \{(x^u_j)_{j=1}^{N^U}\}$, where $x^l_i$/$x^u_j$ denotes the $i^{th}$/$j^{th}$ labeled/unlabeled image, $y^l_i$ is the ground truth for the labeled image, and $N^L$ and $N^U$ ($N^U >> N^L$) are the numbers of labeled and unlabeled samples, respectively.
    \item WSL: The training set $\mathcal{D}$ contains a sparsely annotated labeled set $\mathcal{D}^{WL} = \{(x^l_i, y^{wl}_i)_{i=1}^{N}\}$, where $y^{wl}_i$ is the sparse annotation (\emph{e.g.}, scribble) for the labeled image;
    \item BSL: The training set $\mathcal{D} = \{\mathcal{D}^{WL}, \mathcal{D}^U\}$ comprises a sparsely annotated labeled set $\mathcal{D}^{WL} = \{(x^l_i, y^{wl}_i)_{i=1}^{N^L}\}$ and an unlabeled set $\mathcal{D}^U = \{(x^u_j)_{j=1}^{N^U}\}$;
\end{itemize}

A segmentation model $f(\cdot; \theta)$ is trained on the training set $\mathcal{D}$ and evaluated on an unseen test set $\mathcal{T}$.

\subsection{SynMatch}
This work develops a generic framework, named SynMatch, for medical image segmentation with sparse annotations that involves three settings, \emph{i.e.}, SSL, WSL, and BSL, as defined above.
Figure~\ref{fig:synmatch_framework} illustrates the schematic diagram of SynMatch, which mainly includes two components: \textbf{image synthesis} and \textbf{loss supervision}. The training procedure for SynMatch is summarized in Algorithm~\ref{alg:synmatch_framework}.

\subsubsection{Image synthesis} The main idea of SynMatch is \textit{synthesizing images to match pseudo labels}. 
Its methodology for image synthesis is underpinned by the feature extraction property inherent to convolutional neural networks (CNNs): shallow layers extract low-level \textit{texture} features (e.g., local edges, corners, color), while deep layers encode high-level \textit{shape} information (e.g., anatomical structures)~\cite{hariharan2015hypercolumns,ronneberger2015u,chen2018encoder}. 
SynMatch leverages feature maps extracted from the shallow layer of the segmentation model to capture texture information, and from the deep layers to represent shape. These features are integrated to synthesize complete images without introducing any training parameters.
More importantly, since both the synthesized image and its pseudo label are derived from the same segmentation model for an unlabeled image, they inherently share consistent semantic information, resulting in high consistency between them.
Specifically, the image synthesis procedure consists of two steps: 1) feature extraction, and 2) image generation.

\textit{Feature extraction.} The texture feature $\hat{t}^u_j$ and shape feature $\hat{t}^u_j$ are captured from the segmentation process of an unlabeled image $x^u_j$: 
\begin{equation}
    \hat{t}^u_j, \hat{s}^u_j \gets f(x^u_j, \theta)
\label{eq:feature_extraction}
\end{equation}

\textit{Image generation.} A synthesized image $\hat{x}^s_j$ is generated by fusing these features using a simple weighted combination with a random weight $\alpha \sim U(0, 1)$:
\begin{equation}
    \hat{x}^s_j = \alpha \times \hat{t}^u_j + (1-\alpha) \times \hat{s}^u_j
\label{eq:image generation}
\end{equation}

%-------------------------------------------------------------------------------------------
\newcommand{\PyComment}[1]{\ttfamily{\# #1}}
\begin{algorithm}[!t]
    \caption{SynMatch}
    \label{alg:synmatch_framework}
    \textbf{Input}: {$\mathcal{D}^L = \{(x^l_i, y^{l(w)}_i)_{i=1}^{N^L}\}$, $\mathcal{D}^U = \{(x^u_j)_{j=1}^{N^U}\}$}\\
    \textbf{Parameter}: $\theta$ for the segmentation model \\
    \textbf{Output}: {Trained model $f(\cdot; \theta)$} 
    \begin{algorithmic}[1] 
        \FOR{each iteration}
                \STATE {\PyComment{Pseudo labeling (without gradient) for unlabeled image $x^u_j$ under weak augmentation $\gamma(\cdot)$}}
                \STATE {$\bar{y}^u_j \gets f(\gamma(x^u_j), \theta)$}
                \STATE {\PyComment{Forward propagation for unlabeled/labeled image $x^u_j$/$x^l_i$ under strong augmentation $\Gamma(\cdot)$}}
                \STATE {$\hat{y}^u_j \gets f(\Gamma(x^u_j), \theta)$, $\hat{y}^l_i \gets f(\Gamma(x^l_i), \theta)$}
                \STATE {\PyComment{Image synthesis using texture feature $\hat{t}^u_j$ and shape feature $\hat{s}^u_j$ extracted from the segmentation process of $x^u_j$}}
                \STATE {$\hat{x}^s_j = \alpha \times \hat{t}^u_j + (1-\alpha) \times \hat{s}^u_j$, where $\alpha \sim U(0, 1)$}
                \STATE {\PyComment{Forward propagation for synthetic image $\hat{x}^s_j$}}
                \STATE {$\hat{y}^s_j \gets f(\hat{x}^s_j, \theta)$}
                \STATE {\PyComment{Loss supervision $\mathcal{L} = \mathcal{L}_s + \mathcal{L}_u$ using Eq.~\eqref{eq:ls} and Eq.~\eqref{eq:lu}}}
                \STATE {Back-Propagate $\mathcal{L}$ and Update $\theta$}
        \ENDFOR
        \STATE \textbf{return} $f(\cdot; \theta)$
    \end{algorithmic}
    \footnotesize{Note: since sparsely annotated images contain both labeled and unlabeled pixels, they are considered labeled data for supervised learning and unlabeled data for pseudo supervision. For notational convenience, we use $x^u_j$ uniformly in the algorithm description.}
\end{algorithm}
%-------------------------------------------------------------------------------------------

\subsubsection{Loss supervision}
The loss supervision for SynMatch includes two major components: 
1) the supervised loss $\mathcal{L}_s$ on labeled data and 2) the unsupervised loss $\mathcal{L}_u$ on both the unlabeled and synthesized images, \emph{i.e.}, $\mathcal{L} = \mathcal{L}_s + \mathcal{L}_u$.

\textit{Supervised learning.} For labeled data, the supervised loss $\mathcal{L}_s$ is formulated as:
\begin{equation}
\mathcal{L}_s = \mathcal{L}_{seg}(f(\Gamma(x^l_i), \theta), y^{l(w)}_i)
\label{eq:ls}
\end{equation}
where $y^l_i$ is used when fully-annotated labels are available, \emph{e.g.}, in the semi-supervised learning setting; otherwise, $y^{lw}_i$ is used.
$\mathcal{L}_{seg}$ denotes a segmentation criterion, for which we adopt the joint cross-entropy and Dice loss~\cite{milletari2016v}.

\textit{Unsupervised learning.} The unsupervised loss $\mathcal{L}_u$ comprises two terms on both unlabeled images $x^u_j$ and synthesized images $\hat{x}^s_j$:
\begin{equation}
\begin{aligned}
\mathcal{L}_u & = \underbrace{\mathbb{I}_{\{\bar{p}^u_j \geq \tau\}}(\bar{p}^u_j)\mathcal{L}(f(\Gamma(x^u_j), \theta), \bar{y}^u_j)}_{\mathcal{L}_{org}\text{ on unlabeled image}} \\
&+ \underbrace{\mathbb{I}_{\{\bar{p}^u_j \geq \tau\}}(\bar{p}^u_j)\mathcal{L}(f(\hat{x}^s_j, \theta), \bar{y}^u_j),}_{\mathcal{L}_{syn}\text{ on synthesized image}}
\end{aligned}
\label{eq:lu}
\end{equation}
where $\tau$ is a confidence threshold above which a pixel-level pseudo label is selected for pseudo supervision and $\mathbb{I}(\cdot)$ refers to an indicator function. $\gamma(\cdot)$ represents the weak augmentation that includes image geometric transformations (\emph{e.g.}, cropping, rotation, and flip), while $\Gamma(\cdot)$ denotes the strong augmentation that builds upon $\gamma(\cdot)$ by incorporating pixel intensity transformations (\emph{e.g.}, color jitter and Gaussian blurring) and Mixup operations~\cite{zhang2017mixup,yun2019cutmix}.
Due to the visual differences between synthesized and original images, we apply the unsupervised loss to both unlabeled and synthesized images to prevent the model from overfitting to the synthesized data. Moreover, although each synthesized image is well-aligned with its corresponding pseudo label, we employ a confidence thresholding strategy to ensure that the segmentation targets remain consistent with realistic anatomical structures.

\section{Experiments and Results}
Experiments were conducted on cardiac, skin lesion, and polyp segmentation tasks using the publicly available ACDC~\cite{bernard2018deep}, Kvasir-SEG~\cite{jha2020kvasir}, and ISIC~\cite{codella2019skin,tschandl2018ham10000} datasets under SSL, WSL, and BSL settings.

\subsection{Datasets}
\label{subsec:datasets}
\textbf{ACDC} comprises 200 short-axis cine-MRIs from 100 patients with the corresponding cardiac labels (left ventricle, right ventricle, and myocardium structures). The scribble annotations are provided by~\cite{valvano2021learning}.
Following~\citep{ssl4mis2020}, we split the dataset into training, validation, and test sets according to a ratio of $7:1:2$.

\textbf{ISIC} involves a training set consisting of 2594 dermoscopy images and a test set of 1000 images, along with the corresponding skin lesion annotations. 
We generated scribble annotations from the full annotations using morphology operations\footnote{The scribble annotations will be released along with the code.}.
We split the training set in an $8:2$ ratio, resulting in 2,075 images for training and 519 images for validation. The original test set is used for testing.

\textbf{Kvasir-SEG} contains 1000 colonoscopy images with corresponding polyp annotations. 
Scribble annotations were derived from the full annotations using morphological operations${}^\text{1}$.
We divided the dataset into 700, 100, and 200 images for training, validation, and testing, respectively.

%-------------------------------------------------------------------------
\begin{table}[!t]
\centering
\resizebox{\linewidth}{!}{
\begin{tabular}{l|ll|ll}
\toprule[1pt]
\multirow{3}{*}{Method} & \multicolumn{4}{c}{SSL} \\ \cline{2-5}
& \multicolumn{2}{c|}{5\% $|\mathcal{D}|$} & \multicolumn{2}{c}{10\% $|\mathcal{D}|$}\\
& DSC (\%) $\uparrow$ & ASD $\downarrow$  & DSC (\%) $\uparrow$ & ASD $\downarrow$\\ \midrule[1pt]
BCP (CVPR 2023)${}^\star$  & 87.59  & 0.67  & 88.84  & 1.17 \\
% ABD (+Cross Teaching) (CVPR 2024)*   & 86.35  & 1.22  & 88.52  & 1.43 \\
% ABD (+BCP) (CVPR 2024)*   & 88.96  & 0.52  & 89.81  & 0.49 \\
AD-MT (ECCV 2024)${}^\star$  & 88.75  & \underline{0.50}  & 89.46  & 0.44  \\
CrossMatch (JBHI 2024)${}^\star$  & 88.27  & \textbf{0.46}  & 89.08  & 0.52 \\
GapMatch (AAAI 2025)${}^\star$  & \underline{88.80}  & 0.64  & \underline{90.10}  & \textbf{0.39} \\
W2SPC (MedIA 2025)${}^\star$  & 82.39$\pm.82$  & 2.01$\pm.17$  & 88.92$\pm.64$  & 1.19$\pm.17$\\
FixMatch (NeurIPS 2020) & 83.23$\pm7.67$  & 1.70$\pm2.65$  & 86.26$\pm5.28$  & 0.64$\pm.50$ \\
CPS (CVPR 2021)   & 84.76$\pm5.91$  & .93$\pm.80$  & 86.76$\pm5.30$  & 0.70$\pm.50$ \\
UniMatch (CVPR 2023)  & 87.27$\pm4.97$  & 0.59$\pm.34$  & 88.23$\pm5.45$  & 0.58$\pm.36$\\
SynMatch (ours)  & \textbf{88.96}$\pm4.14$  & \textbf{0.46}$\pm.26$  & \textbf{90.18}$\pm3.54$  & \underline{0.40}$\pm.27$\\
\midrule[1pt]
\midrule[1pt]
\multirow{3}{*}{Method} & \multicolumn{4}{c}{WSL} \\ \cline{2-5}
& \multicolumn{4}{c}{100\% $|\mathcal{D}|$ (scribbles)} \\
& \multicolumn{2}{c}{DSC (\%) $\uparrow$} & \multicolumn{2}{c}{ASD $\downarrow$} \\ \midrule[1pt]
DMPL (MICCAI 2022)${}^\star$ & \multicolumn{2}{c|}{\underline{87.20}$\pm7.70$}  & \multicolumn{2}{c}{-} \\
CycleMix (CVPR 2022)  & \multicolumn{2}{c|}{85.23$\pm4.53$}  & \multicolumn{2}{c}{\underline{0.73}$\pm.79$} \\
% ScribbleVC (ACM MM 2023)  \\
% DMSPS (MedIA 2024)  \\
FixMatch (NeurIPS 2020) & \multicolumn{2}{c|}{81.19$\pm5.27$}  & \multicolumn{2}{c}{3.11$\pm2.74$} \\
CPS (CVPR 2021)  & \multicolumn{2}{c|}{85.38$\pm4.14$}  & \multicolumn{2}{c}{0.57$\pm.96$} \\
UniMatch (CVPR 2023) & \multicolumn{2}{c|}{86.24$\pm4.42$}  & \multicolumn{2}{c}{0.76$\pm1.15$} \\
GapMatch (AAAI 2025)  & \multicolumn{2}{c|}{85.39$\pm5.21$}  & \multicolumn{2}{c}{1.10$\pm2.38$} \\
SynMatch (ours) & \multicolumn{2}{c|}{\textbf{87.93}$\pm3.68$}  & \multicolumn{2}{c}{\textbf{0.55}$\pm.60$${}^\ast$} \\
\midrule[1pt]
\midrule[1pt]
\multirow{3}{*}{Method} & \multicolumn{4}{c}{BSL} \\ \cline{2-5}
& \multicolumn{2}{c|}{5\% $|\mathcal{D}|$ (scribbles)} & \multicolumn{2}{c}{10\% $|\mathcal{D}|$ (scribbles)}\\
& DSC (\%) $\uparrow$ & ASD $\downarrow$  & DSC (\%) $\uparrow$ & ASD $\downarrow$\\ \midrule[1pt]
FixMatch (NeurIPS 2020)  & 54.18$\pm26.60$  & 11.78$\pm22.11$  & 74.95$\pm10.12$  & 2.11$\pm3.35$\\
CPS (CVPR 2021)  & 55.45$\pm27.96$  & 10.19$\pm16.90$  & 76.56$\pm8.93$  & 0.93$\pm.80$ \\
CycleMix (CVPR 2022)  & 75.68$\pm9.88$  & 2.82$\pm3.25$  & 79.64$\pm6.79$  & 2.07$\pm2.77$ \\
UniMatch (CVPR 2023) & \underline{80.68}$\pm5.46$  & \underline{1.58}$\pm.63$  & \underline{84.42}$\pm5.15$  & \underline{0.83}$\pm1.04$ \\
GapMatch (AAAI 2025)  & 75.42$\pm13.35$  & 3.00$\pm3.93$  & 79.07$\pm6.55$  & 1.76$\pm2.57$ \\
SynMatch (ours) & \textbf{84.15}$\pm4.62$${}^\ast$  & \textbf{0.58}$\pm.28$${}^\ast$  & \textbf{86.67}$\pm4.11$${}^\ast$  & \textbf{0.50}$\pm.24$${}^\ast$\\
\midrule[1pt]
\midrule[1pt]
Full Supervision & \multicolumn{2}{c}{DSC: 89.83$\pm4.28$}   & \multicolumn{2}{c}{ASD: 0.64$\pm.64$}\\
\bottomrule[1pt]
\end{tabular}
}
\caption{Comparison with state-of-the-art methods on the cardiac segmentation task under SSL, WSL, and BSL settings.
${}^\star$ Results reported in the original paper. \#\% $|\mathcal{D}|$ (scribbles) denotes the training set consisting of \#\% sparsely labeled data with scribble annotations and and 1 - \#\% unlabeled data.
Standard deviation is calculated over samples.
The best and second-best results are highlighted in \textbf{bold} and \underline{underline}, respectively.
${}^\ast$ indicates statistically significant improvements ($p$-value $ < 0.05$ based on a paired t-test) of SynMatch over the second-best results (Note: this is \textit{NA} for second-best results adopted from their original papers).}
\label{Tab:acdc}
\end{table}
%-------------------------------------------------------------------------

\subsection{Implementation Details}
\textbf{Hardware:} NVIDIA A40 GPU with 48GB GPU memory.
\textbf{Software:} Python 3.8, PyTorch~\citep{paszke2019pytorch} 1.11.0, CUDA 11.3.
\textbf{Optimizer:} AdamW~\citep{kingma2014adam} with a fixed learning rate of 1e-4, and the training time is set to 200 epochs for all experiments.
\textbf{Framework:} U-Net~\citep{ronneberger2015u} is used as the segmentation backbone.
We leverage the feature maps after the first convolutional block of U-Net as texture features and those before the final convolutional block (before the segmentation head) as shape features for image synthesis. For the three-channel colonoscopy and dermoscopy images, SynMatch first synthesizes the luminance channel and then merges it with the original image’s chrominance channels to generate the synthesized image.
Following FixMatch~\citep{sohn2020fixmatch}, we set the confidence threshold $\tau = 0.95$.
\textbf{Data:} images are resized to $256 \times 256$ for training and testing, while the predictions are recovered to their original sizes for performance evaluation. 
We perform 2D slice segmentation rather than 3D volume segmentation on the ACDC dataset due to its large slice thickness, and the resulting predictions are then stacked into a 3D volume for evaluation.
\textbf{Evaluation metrics:} Dice similarity coefficient (DSC) and average surface distance (ASD).

\subsection{Comparison with SOTAs}
We extensively compared SynMatch with advanced semi-supervised models in recent years, including BCP~\cite{bai2023bidirectional}, AD-MT~\cite{zhao2024alternate}, CrossMatch~\cite{zhao2024crossmatch}, W2SPC~\cite{yang2025semi}, FixMatch~\cite{sohn2020fixmatch}, CPS~\cite{chen2021semi},  UniMatch~\cite{yang2023revisiting} as well as GapMatch~\cite{huang2025gapmatch}, and weak-supervised models DMPL~\cite{luo2022scribble} as well as CycleMix~\cite{zhang2022cyclemix}, under the increasing complexity segmentation tasks (cardiac, skin lesion and polyp) and sparse annotation scenarios (SSL, WSL and BSL), respectively. 
In a nutshell, SynMatch set state-of-the-art performance on cardiac segmentation [Table~\ref{Tab:acdc}], skin lesion segmentation [Table~\ref{Tab:isic}], and polyp segmentation [Table~\ref{Tab:kvasir}] tasks under the three settings.

%-------------------------------------------------------------------------
\begin{table}[!t]
\centering
\resizebox{\linewidth}{!}{
\begin{tabular}{l|ll|ll}
\toprule[1pt]
\multirow{3}{*}{Method} & \multicolumn{4}{c}{SSL} \\ \cline{2-5}
& \multicolumn{2}{c|}{5\% $|\mathcal{D}|$} & \multicolumn{2}{c}{10\% $|\mathcal{D}|$}\\
& DSC (\%) $\uparrow$ & ASD $\downarrow$  & DSC (\%) $\uparrow$ & ASD $\downarrow$\\ \midrule[1pt]
FixMatch (NeurIPS 2020)  & 85.20$\pm15.07$  & 1.83$\pm5.97$  & 85.39$\pm13.67$  & 1.74$\pm5.19$ \\
CPS (CVPR 2021) & 84.82$\pm14.11$  & 2.40$\pm6.11$  & 85.99$\pm13.46$  & \textbf{1.54}$\pm4.96$ \\
UniMatch (CVPR 2023) & 85.06$\pm14.17$  & 2.01$\pm5.84$  & \underline{86.74}$\pm13.54$  & 1.61$\pm5.04$ \\
GapMatch (AAAI 2025)  & \underline{85.37}$\pm13.95$  & \underline{1.76}$\pm5.27$  & 85.49$\pm13.82$  & 1.75$\pm5.16$ \\
SynMatch (ours)  & \textbf{85.53}$\pm14.23$${}^\ast$  & \textbf{1.63}$\pm5.02$${}^\ast$ & \textbf{86.78}$\pm13.87$  & \underline{1.55}$\pm5.22$   \\
\midrule[1pt]
\midrule[1pt]
\multirow{3}{*}{Method} & \multicolumn{4}{c}{WSL} \\ \cline{2-5}
& \multicolumn{4}{c}{100\% $|\mathcal{D}|$ (scribbles)} \\
& \multicolumn{2}{c}{DSC (\%) $\uparrow$} & \multicolumn{2}{c}{ASD $\downarrow$}  \\ \midrule[1pt]
FixMatch (NeurIPS 2020)  & \multicolumn{2}{c|}{76.49$\pm18.91$}  & \multicolumn{2}{c}{\underline{2.73}$\pm12.08$}\\
CPS (CVPR 2021) & \multicolumn{2}{c|}{76.34$\pm20.62$}  & \multicolumn{2}{c}{2.73$\pm12.86$} \\
CycleMix (CVPR 2022)  & \multicolumn{2}{c|}{\underline{77.76}$\pm16.57$}  & \multicolumn{2}{c}{2.79$\pm4.52$} \\
UniMatch (CVPR 2023)  & \multicolumn{2}{c|}{77.65$\pm17.97$}  & \multicolumn{2}{c}{2.26$\pm4.01$} \\
GapMatch (AAAI 2025)   & \multicolumn{2}{c|}{77.68$\pm16.49$}  & \multicolumn{2}{c}{2.86$\pm4.82$} \\
SynMatch (ours) & \multicolumn{2}{c|}{\textbf{79.24}$\pm16.98$${}^\ast$}  & \multicolumn{2}{c}{\textbf{2.19}$\pm3.18$${}^\ast$} \\
\midrule[1pt]
\midrule[1pt]
\multirow{3}{*}{Method} & \multicolumn{4}{c}{BSL} \\ \cline{2-5}
& \multicolumn{2}{c|}{5\% $|\mathcal{D}|$ (scribbles)} & \multicolumn{2}{c}{10\% $|\mathcal{D}|$ (scribbles)}\\
& DSC (\%) $\uparrow$ & ASD $\downarrow$  & DSC (\%) $\uparrow$ & ASD $\downarrow$\\ \midrule[1pt]
FixMatch (NeurIPS 2020)   & 76.58$\pm19.79$  & 2.28$\pm11.04$  & 75.04$\pm21.32$  & 2.83$\pm12.83$  \\
CPS (CVPR 2021) & 75.77$\pm22.08$  & 2.88$\pm13.89$  & 73.02$\pm22.45$  & 3.12$\pm13.75$ \\
CycleMix (CVPR 2022)  & 74.87$\pm16.42$  & 3.26$\pm6.53$  & 74.74$\pm16.56$  & 3.14$\pm6.00$ \\
UniMatch (CVPR 2023) & \underline{76.88}$\pm18.22$  & \textbf{1.94}$\pm4.70$  & \underline{77.59}$\pm18.71$  & \underline{2.36}$\pm8.84$ \\
GapMatch (AAAI 2025)  & 76.38$\pm18.83$  & \underline{1.97}$\pm5.14$  & 73.58$\pm21.88$  & 2.93$\pm13.26$ \\
SynMatch (ours)  & \textbf{77.34}$\pm17.06$${}^\ast$  & 2.31$\pm5.29$  & \textbf{78.42}$\pm17.76$${}^\ast$  & \textbf{2.08}$\pm7.17$${}^\ast$\\
\midrule[1pt]
\midrule[1pt]
Full Supervision & \multicolumn{2}{c}{DSC: 87.38$\pm12.57$}   & \multicolumn{2}{c}{ASD: 1.32$\pm3.83$}\\
\bottomrule[1pt]
\end{tabular}
}
\caption{Comparison with state-of-the-art methods on the skin lesion segmentation task under SSL, WSL, and BSL settings.
\#\% $|\mathcal{D}|$ (scribbles) denotes the training set consisting of \#\% sparsely labeled data with scribble annotations and and 1 - \#\% unlabeled data.
Standard deviation is calculated over samples.
The best and second-best results are highlighted in \textbf{bold} and \underline{underline}, respectively.
${}^\ast$ indicates statistically significant improvements ($p$-value $ < 0.05$ based on a paired t-test) of SynMatch over the second-best results.}
\label{Tab:isic}
\end{table}
%-------------------------------------------------------------------------

%-------------------------------------------------------------------------
\begin{table}[!t]
\centering
\resizebox{\linewidth}{!}{
\begin{tabular}{l|ll|ll}
\toprule[1pt]
\multirow{3}{*}{Method} & \multicolumn{4}{c}{SSL} \\ \cline{2-5}
& \multicolumn{2}{c|}{5\% $|\mathcal{D}|$} & \multicolumn{2}{c}{10\% $|\mathcal{D}|$}\\
& DSC (\%) $\uparrow$ & ASD $\downarrow$  & DSC (\%) $\uparrow$ & ASD $\downarrow$\\ \midrule[1pt]
FixMatch (NeurIPS 2020) & 59.62$\pm28.00$  & 12.07$\pm25.76$  & 61.97$\pm28.02$  & 9.61$\pm21.31$ \\
CPS (CVPR 2021)  & 65.15$\pm28.01$  & 10.23$\pm27.69$  & 68.84$\pm26.19$  & \textbf{7.52}$\pm22.96$\\
UniMatch (CVPR 2023)  & \underline{68.70}$\pm25.35$  & \underline{9.41}$\pm29.95$  & \underline{72.65}$\pm26.07$  & 9.84$\pm31.46$ \\
GapMatch (AAAI 2025)  & 65.62$\pm28.53$  & 9.54$\pm25.27$  & 68.06$\pm27.52$  & 9.23$\pm19.80$ \\
SynMatch (ours)  & \textbf{69.38}$\pm27.79$${}^\ast$  & \textbf{8.50}$\pm25.49$${}^\ast$  & \textbf{72.92}$\pm26.61$${}^\ast$  & \underline{9.11}$\pm30.17$ \\
\midrule[1pt]
\midrule[1pt]
\multirow{3}{*}{Method} & \multicolumn{4}{c}{WSL} \\ \cline{2-5}
& \multicolumn{4}{c}{100\% $|\mathcal{D}|$ (scribbles)} \\
& \multicolumn{2}{c}{DSC (\%) $\uparrow$} & \multicolumn{2}{c}{ASD $\downarrow$}  \\ \midrule[1pt]
FixMatch (NeurIPS 2020)  & \multicolumn{2}{c|}{68.70$\pm21.27$}  & \multicolumn{2}{c}{10.31$\pm16.94$}\\
CPS (CVPR 2021) & \multicolumn{2}{c|}{69.93$\pm21.61$}  & \multicolumn{2}{c}{10.00$\pm17.69$}\\
CycleMix (CVPR 2022)  & \multicolumn{2}{c|}{66.00$\pm20.40$}  & \multicolumn{2}{c}{12.08$\pm17.68$} \\
UniMatch (CVPR 2023) & \multicolumn{2}{c|}{\underline{72.69}$\pm20.98$}  & \multicolumn{2}{c}{\underline{6.66}$\pm15.93$} \\
GapMatch (AAAI 2025)  & \multicolumn{2}{c|}{68.42$\pm21.56$}  & \multicolumn{2}{c}{11.30$\pm17.60$} \\
SynMatch (ours)  & \multicolumn{2}{c|}{\textbf{73.89}$\pm20.69$${}^\ast$}  & \multicolumn{2}{c}{\textbf{6.64}$\pm17.01$}\\
\midrule[1pt]
\midrule[1pt]
\multirow{3}{*}{Method} & \multicolumn{4}{c}{BSL} \\ \cline{2-5}
& \multicolumn{2}{c|}{5\% $|\mathcal{D}|$ (scribbles)} & \multicolumn{2}{c}{10\% $|\mathcal{D}|$ (scribbles)}\\
& DSC (\%) $\uparrow$ & ASD $\downarrow$  & DSC (\%) $\uparrow$ & ASD $\downarrow$\\ \midrule[1pt]
FixMatch (NeurIPS 2020) & 39.02$\pm15.52$  & 14.83$\pm23.83$  & 56.20$\pm24.58$  & 18.44$\pm28.72$ \\
CPS (CVPR 2021)  & 31.88$\pm15.00$  & 23.76$\pm31.14$  & 60.81$\pm24.97$  & 10.35$\pm19.96$ \\
CycleMix (CVPR 2022)  & 48.37$\pm21.58$  & 14.64$\pm21.78$  & 55.63$\pm23.18$  & 16.27$\pm21.60$ \\
UniMatch (CVPR 2023)  & \underline{62.93}$\pm22.54$  & \underline{10.33}$\pm22.63$  & \underline{68.21}$\pm23.37$  & \underline{8.50}$\pm22.67$ \\
GapMatch (AAAI 2025)  & 35.43$\pm14.24$  & 27.61$\pm26.67$  & 59.31$\pm24.09$  & 16.11$\pm28.49$ \\
SynMatch (ours)   & \textbf{65.14}$\pm24.18$${}^\ast$  & \textbf{9.32}$\pm18.50$${}^\ast$   & \textbf{69.36}$\pm23.45$${}^\ast$  & \textbf{8.43}$\pm19.03$${}^\ast$\\
\midrule[1pt]
\midrule[1pt]
Full Supervision & \multicolumn{2}{c}{78.53$\pm22.89$}   & \multicolumn{2}{c}{5.87$\pm16.50$}\\
\bottomrule[1pt]
\end{tabular}
}
\caption{Comparison with state-of-the-art methods on the polyp segmentation task under SSL, WSL, and BSL settings.
\#\% $|\mathcal{D}|$ (scribbles) denotes the training set consisting of \#\% sparsely labeled data with scribble annotations and and 1 - \#\% unlabeled data.
Standard deviation is calculated over samples.
The best and second-best results are highlighted in \textbf{bold} and \underline{underline}, respectively.
${}^\ast$ indicates statistically significant improvements ($p$-value $ < 0.05$ based on a paired t-test) of SynMatch over the second-best results.
}
\label{Tab:kvasir}
\end{table}
%-------------------------------------------------------------------------

%-------------------------------------------------------------------------------------------
\begin{table*}[!t]
\centering
\resizebox{\linewidth}{!}{
\begin{tabular}{cc|cc|cc|cc|cc|cc}
\toprule[1pt]
\multicolumn{2}{c|}{SynMatch}  & \multicolumn{4}{c|}{Semi-Supervised Learning}  & \multicolumn{2}{c|}{Weakly-Supervised Learning}  & \multicolumn{4}{c}{Barely-Supervised Learning}  \\ \midrule
\multirow{2}{*}{Strong-Weak Supervision ($\mathcal{L}_{org}$)} 
& \multirow{2}{*}{Synthesis Supervision ($\mathcal{L}_{syn}$)}  
& \multicolumn{2}{c|}{5\% $|\mathcal{D}|$} & \multicolumn{2}{c|}{10\% $|\mathcal{D}|$} & \multicolumn{2}{c|}{100\% $|\mathcal{D}|$ (scribbles)} & \multicolumn{2}{c|}{5\% $|\mathcal{D}|$ (scribbles)} & \multicolumn{2}{c}{10\% $|\mathcal{D}|$ (scribbles)} \\
&  & DSC  $\uparrow$ & ASD $\downarrow$ & DSC $\uparrow$ & ASD $\downarrow$ & DSC  $\uparrow$ & ASD $\downarrow$ & DSC  $\uparrow$ & ASD $\downarrow$ & DSC $\uparrow$ & ASD $\downarrow$\\
\midrule[1pt]
&      & 71.97$\pm14.94$ & 2.71$\pm3.43$ & 84.05$\pm7.25$ & 1.30$\pm1.55$ & 80.48$\pm4.89$ & 3.88$\pm3.59$         & 26.71$\pm8.39$ &  62.23$\pm8.29$  & 74.82$\pm11.76$ &  3.45$\pm3.57$\\
$\surd$ & & 83.23$\pm7.67$  & 1.70$\pm2.65$  & 86.26$\pm5.28$  & 0.64$\pm.50$ & 81.19$\pm5.27$  & 3.11$\pm2.74$ & 54.18$\pm26.60$ &  11.78$\pm22.11$  & 74.95$\pm10.12$ &  2.11$\pm3.35$ \\
& $\surd$  & 85.28$\pm6.21$  & 1.14$\pm1.91$ & 87.95$\pm3.88$  & 0.63$\pm.60$ & 85.71$\pm5.14$  & 0.81$\pm2.05$ & 79.88$\pm8.46$ &  0.90$\pm.89$  & 82.57$\pm7.21$ &  0.99$\pm1.41$ \\
$\surd$ & $\surd$ & \textbf{88.96}$\pm4.14$  & \textbf{0.46}$\pm.26$  & \textbf{90.18}$\pm3.54$  & \underline{0.40}$\pm.27$  & \textbf{87.93}$\pm3.68$  & \textbf{0.55}$\pm.60$ & \textbf{84.15}$\pm4.62$ &  \textbf{0.58}$\pm.28$  & \textbf{86.67}$\pm4.11$ &  \textbf{0.50}$\pm.24$ \\
\bottomrule[1pt]
\end{tabular}
}
\caption{Ablation study for SynMatch on the cardiac segmentation task under SSL, BSL, and SSL settings.
Standard deviation is calculated over samples.
The best results are highlighted in \textbf{bold}.
}
\label{Tab:ablation}
\end{table*}
%--------------------------------------------------

\subsubsection{Results on Cardiac MRI}
We first evaluated SynMatch on the cardiac segmentation task with more regular target shapes.
As reported in Table~\ref{Tab:acdc}, SynMatch outperforms all compared methods under the SSL, WSL, and BSL settings. 
While the performance gain under the SSL setting is relatively modest, it becomes more substantial in the most challenging BSL scenario with extremely sparse annotations.
For instance, the DSC improvement over UniMatch increases from 1.69\% in the SSL setting to 3.47\% in the BSL setting, while the improvement over GapMatch grows from 0.16\% to 8.83\% across the two scenarios.

\subsubsection{Results on Dermoscopy}
We then conducted further evaluations of SynMatch on the skin lesion segmentation task with irregular targets in dermoscopy images.
In Table~\ref{Tab:isic}, all methods under the SSL setting achieve satisfactory performance, approaching that of full supervision (i.e., U-Net trained with 100\% labeled data) due to the large-scale training set.
However, these methods experience performance degradation under the BSL setting, primarily due to the accumulation of confirmation bias caused by inconsistent unlabeled-image and pseudo-label pairs.
In contrast, SynMatch achieves the best performance, with DSC of 77.34\% and 78.42\% under 5\% and 10\% scribble-annotated labeled data, respectively.

\subsubsection{Results on Colonoscopy}
Finally, we tested SynMatch on the polyp segmentation task, which is more challenging due to heterogeneous lesions and relatively small-scale training data. 
As shown in Table~\ref{Tab:kvasir}, all methods perform better under the WSL setting than in the SSL setting, as scribble sparse annotations are likely to cover a broader range of heterogeneous lesions, whereas dataset-level sparse labels in the SSL setting are typically limited to a narrower subset of lesion types.
On the other hand, SynMatch consistently outperforms the compared methods, with substantial performance gains in the BSL setting.
For example, it achieves a 3.45\% relative improvement in DSC compared with the second-best method under the BSL setting with 5\% scribble-annotated labeled data. 
Moreover, SynMatch exhibits a relatively small performance drop from SSL to BSL, \emph{e.g.}, under 5\% labeled data, an absolute DSC degradation of only 4.24\% for SynMatch compared with 30.19\% for GapMatch, a recent strong-weak pseudo supervision-based method.

\subsubsection{Summary} 
The quantitative results shown above highlight several important insights. 
Strong-weak pseudo supervision-based methods struggle to generate highly consistent unlabeled image–pseudo-label pairs and lack mechanisms to recognize and correct their errors. Consequently, the segmentation models are inclined to memorize and accumulate these errors, ultimately degrading the effectiveness of consistency training. 
This issue becomes more crucial under the BSL setting, where segmentation models often generate noiser pseudo labels due to the extremely limited annotations. 
In contrast, our SynMatch outperforms the competing methods by a large margin in the BSL scenario. For example, significant performance improvements are observed on the polyp segmentation task [Table~\ref{Tab:kvasir}].
Overall, these results corroborate the versatility and robustness of SynMatch across multiple tasks under varying levels of supervision.

\subsubsection{Qualitative Results} 
\begin{figure}[t]
\centering
\includegraphics[width=\linewidth]{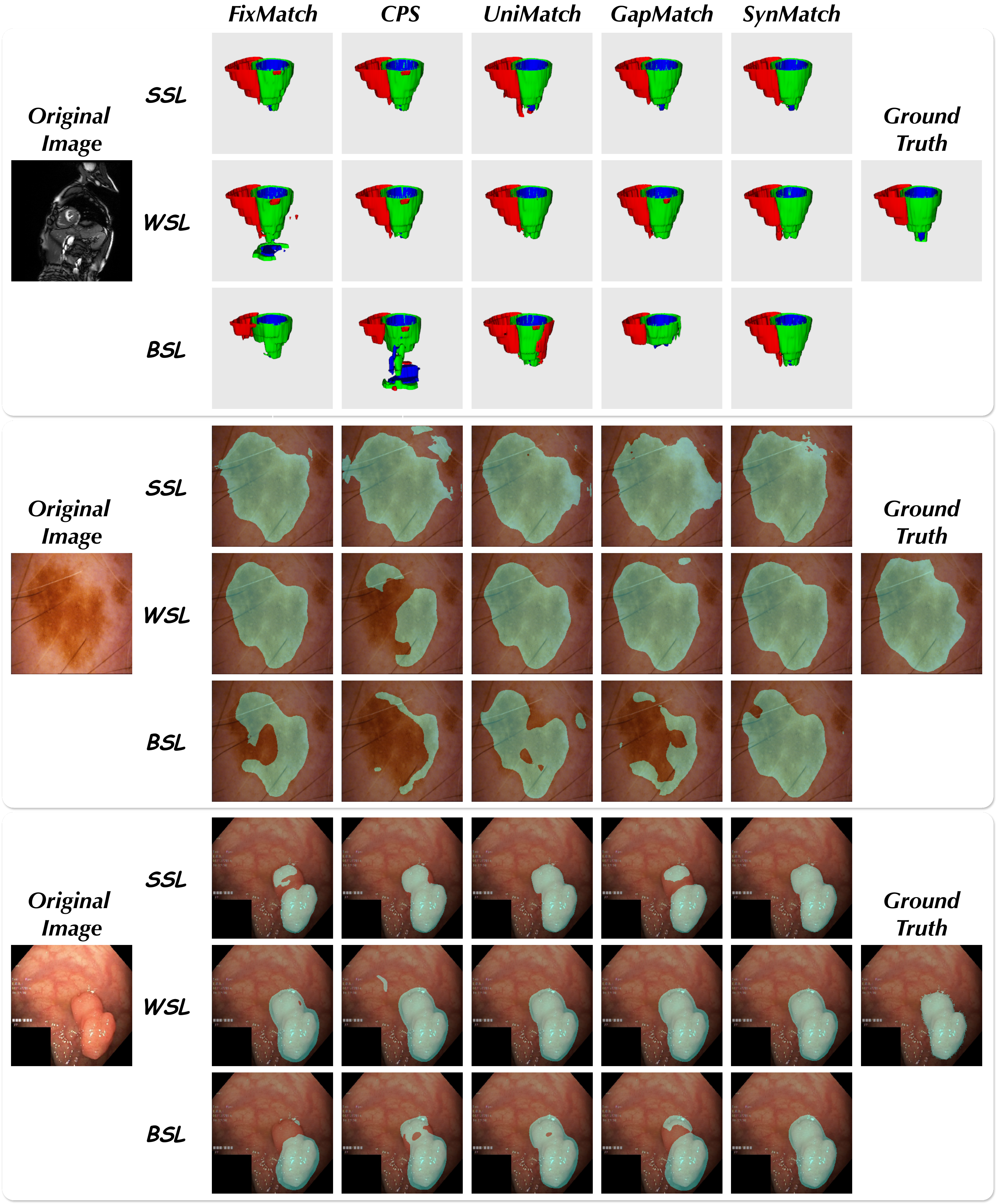} 
\caption{Segmentation examples on the ACDC, ISIC, and Kvasir-SEG dataset under SSL, WSL, and BSL settings.}
\label{fig:qualitative_example}
\end{figure}

Figure~\ref{fig:qualitative_example} provides some segmentation examples on the ACDC, ISIC, and Kvasir-SEG datasets under the SSL, WSL, and BSL settings, respectively. 
Specifically, for the compared methods, the limited labeled samples cannot capture the complexity and variability of lesions, resulting in a loss of fine-grained details, especially under the BSL setting, where numerous regions are incorrectly segmented.
In contrast, due to its training on highly consistent synthesized-image and pseudo-label pairs, SynMatch produces more accurate segmentation maps in delineating organs and lesions compared with other methods.
These qualitative results align with the quantitative performance reported in the previous section.

%--------------------------------------------------------------------------------------------------------------
%--------------------------------------------------------------------------------------------------------------

\subsection{Ablation Study}
We investigated the effectiveness of each component of SynMatch [Table~\ref{Tab:ablation}] on the ACDC dataset under SSL, WSL, and BSL settings.
Moreover, we analyzed the impact of different feature fusion strategies [Figure~\ref{fig:feature_fusion}] and validated the high consistency between synthesized images and pseudo labels both quantitatively [Figure~\ref{fig:synmatch_consistency}] and qualitatively [Figure~\ref{fig:synmatch_synthesized_image}] using the ACDC dataset under the more challenging BSL setting with 5\% and 10\% labeled data (scribbles).

\subsubsection{Effectiveness of Each Component}
Table~\ref{Tab:ablation} shows progressively performance improvements as the strong-weak pseudo supervision $\mathcal{L}_{org}$ and synthesized image supervision $\mathcal{L}_{syn}$ are gradually incorporated into the supervised segmentation baseline, \emph{i.e.}, U-Net. 
These gains can be attributed to the additional supervisory signals provided by both unlabeled and synthesized images during training.
Notably, due to the high consistency between synthesized images and their corresponding pseudo labels, the synthesis supervision yields greater performance improvements compared with the strong-weak supervision counterpart.
Although the synthesis supervision branch enables SynMatch to alleviate the issue of confirmation bias, relying solely on the synthesis supervision leads to overfitting to the synthesized data.
Finally, the combination of strong-weak and synthesis supervision achieves the best overall performance, demonstrating the complementarity of the two supervisions.

\subsubsection{Analysis of Feature Fusion} 
\begin{figure}[t]
\centering
\includegraphics[width=.85\linewidth]{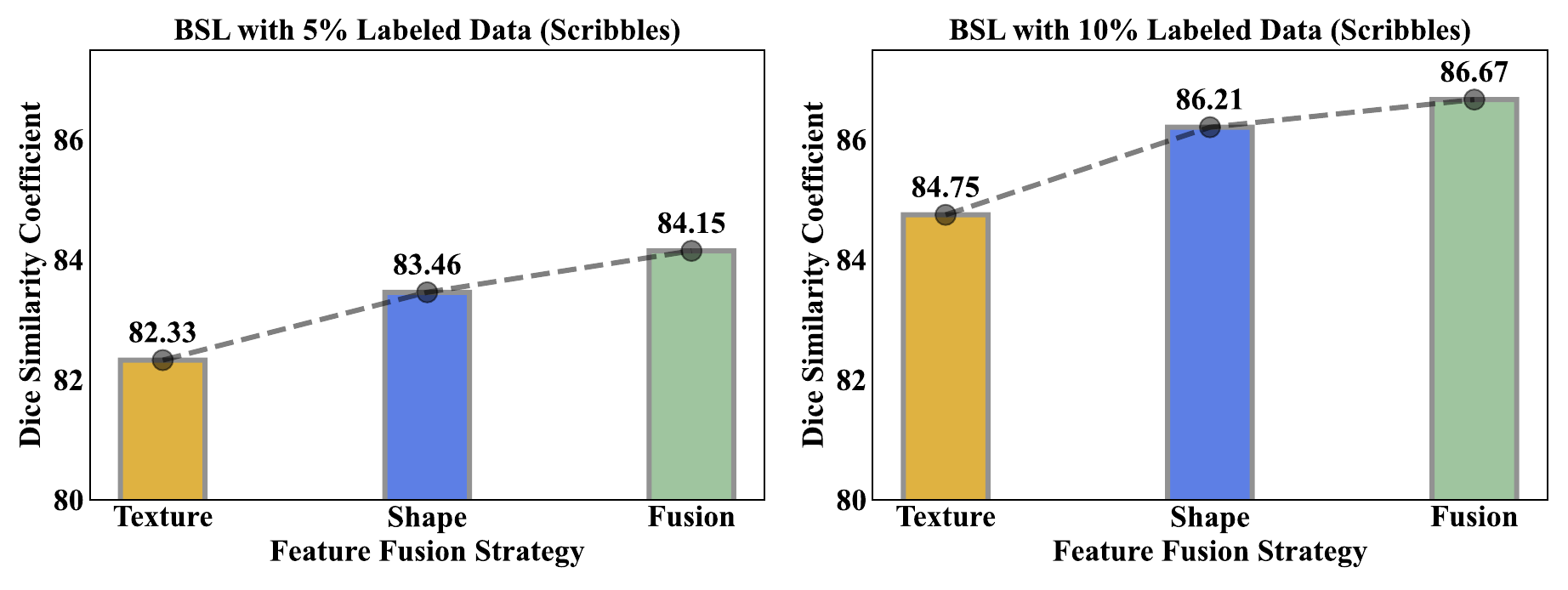} 
\caption{Performance of SynMatch using different features for image synthesis on the ACDC dataset under the BSL setting with 5\% and 10\% labeled data (scribbles).}
\label{fig:feature_fusion}
\end{figure}

Figure~\ref{fig:feature_fusion} presents the segmentation performance of SynMatch using three different types of features for image synthesis on the ACDC dataset under the BSL setting with 5\% and 10\% labeled data (scribbles): 1) texture features, 2) shape features, and 3) a weighted combination of texture and shape features, where the weighted combination strategy achieves the best performance.
This result can be attributed to the integration of texture and shape features, which ensures that the synthesized images retain comprehensive image properties similar to those of real images. In contrast, images generated using texture features alone tend to exhibit shape inconsistencies with their corresponding pseudo labels. On the other hand, images synthesized using only shape features, while highly consistent with pseudo labels, may introduce shortcuts that impair the segmentation model’s generalization to real images.

%--------------------------------------------------------------------------------------------------------------
%--------------------------------------------------------------------------------------------------------------

\subsubsection{Analysis of Semantic Consistency}
\begin{figure}[!t]
\centering
\includegraphics[width=.85\linewidth]{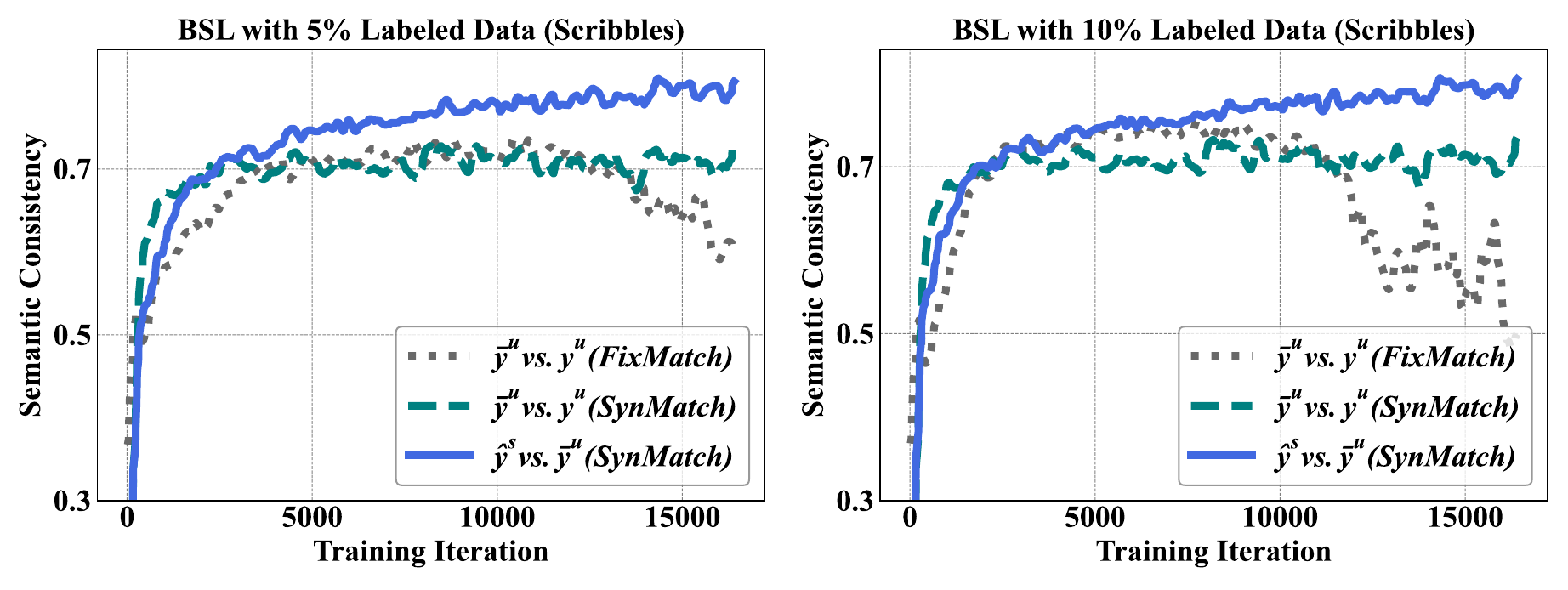} 
\caption{Semantic Consistency of FixMatch and SynMatch on the ACDC dataset under the BSL setting with 5\% and 10\% labeled data (scribbles). 
}
\label{fig:synmatch_consistency}
\end{figure}

\begin{figure}[!t]
\centering
\includegraphics[width=\linewidth]{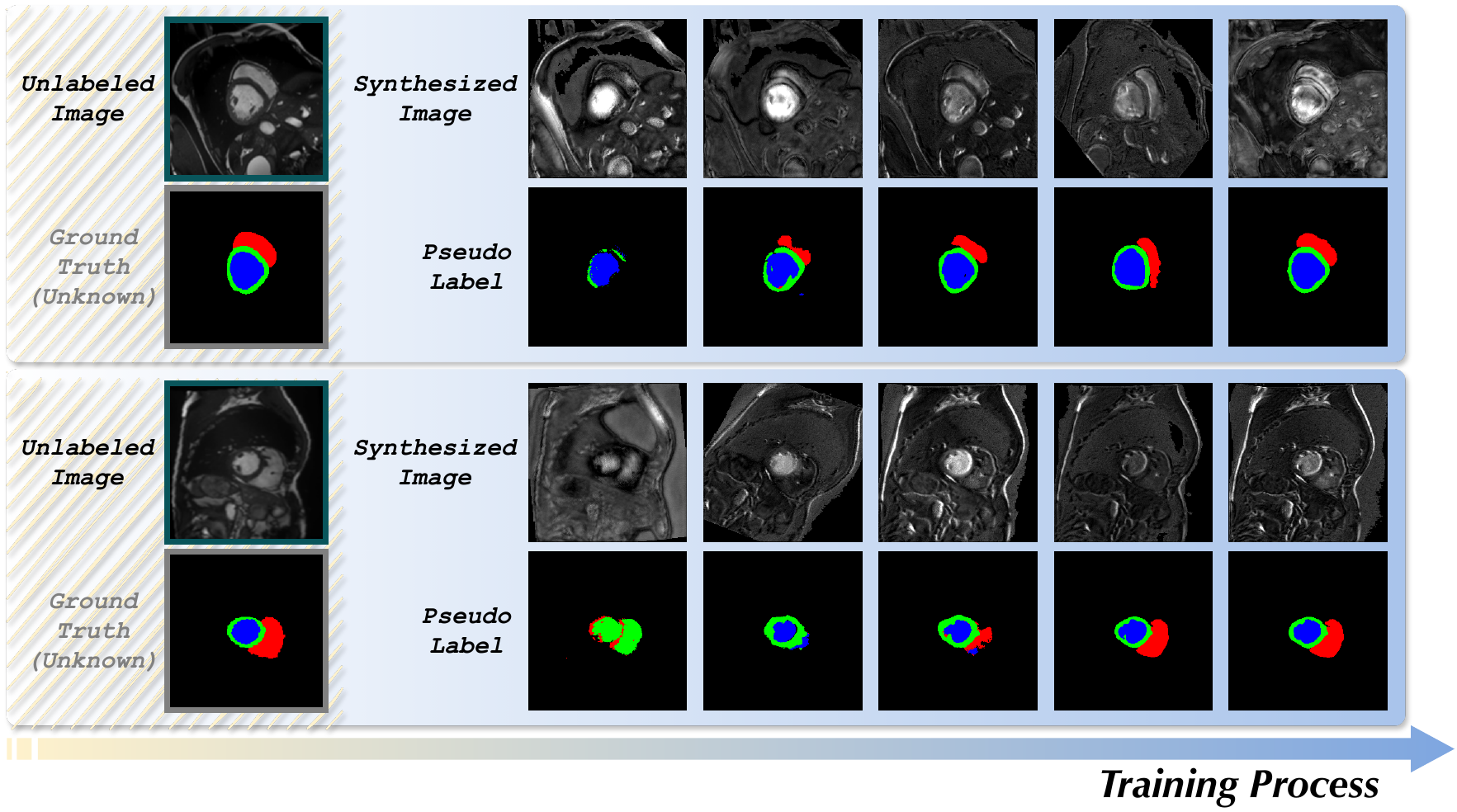} 
\caption{Visualization of SynMatch synthesized unlabeled-image and pseudo-label pairs during the training process.}
\label{fig:synmatch_synthesized_image}
\end{figure}

Figure~\ref{fig:synmatch_consistency} depicts the semantic consistency of FixMatch and SynMatch.
Specifically, in FixMatch (with only strong-weak pseudo supervision on original unlabeled data), pseudo label errors tend to accumulate, which leads to lower consistency between the pseudo label and the ground truth, \textit{i.e.}, $\bar{y}^u \text{ \textit{vs.} } y^u$, thereby resulting in lower segmentation performance.
In contrast, SynMatch achieves noticeably higher consistency for both `$\hat{y}^s \text{ \textit{vs.} } \bar{y}^u$' and `$\bar{y}^u \text{ \textit{vs.} } y^u$' (\textit{i.e.}, segmentation ability).
These results suggest that training with highly consistent synthesized image–pseudo label pairs facilitates the learning of the model and enhances the model's ability to generate high-quality pseudo labels for unlabeled images.
It also further verifies the correlation between semantic consistency during training and segmentation performance.
Moreover, we provide some qualitative examples of synthesized images by SynMatch in Figure~\ref{fig:synmatch_synthesized_image}. We can observe that the synthesized images align well with the pseudo labels, and the quality of pseudo labels is gradually improved during training.

%--------------------------------------------------------------------------------------------------------------
%--------------------------------------------------------------------------------------------------------------

%--------------------------------------------------------------------------------------------------------------
%--------------------------------------------------------------------------------------------------------------

\section{Conclusion}
We propose SynMatch, a novel framework for medical image segmentation with sparse annotations. 
The key idea of SynMatch lies in synthesizing images to match the pseudo labels.
We validate the effectiveness of this idea through extensive experiments on cardiac, skin lesion, and polyp segmentation tasks under SSL, WSL, and BSL settings. The results demonstrate that SynMatch is a simple yet effective method with strong potential as a generic framework.
In addition, we attribute the failure of the strong-weak pseudo supervision paradigm in the BSL setting to the accumulation of inaccurate pseudo labels, \emph{i.e.}, confirmation bias.
Future work could explore more advanced feature fusion and image synthesis techniques to enhance the quality of synthesized images, as well as incorporate more powerful segmentation backbones to boost segmentation performance. Moreover, the three settings with varying degrees of label scarcity on each dataset can serve as a new benchmark for advancing research in label-efficient medical image segmentation.

\bibliography{aaai2026}

\end{document}